\documentclass{article}

\usepackage[sglblindworkshop, final]{neurips_2025}
\workshoptitle{Learning from Time Series for Health}

\usepackage[utf8]{inputenc} 
\usepackage[T1]{fontenc}    
\usepackage{hyperref}       
\usepackage{url}            
\usepackage{booktabs}       
\usepackage{amsfonts}       
\usepackage{nicefrac}       
\usepackage{microtype}      
\usepackage{xcolor}         
\usepackage{amsmath}
\usepackage{amssymb, amsmath}

\usepackage{graphicx}
\usepackage{subcaption}
\usepackage{siunitx}
\usepackage{amsthm}

\title{Learning Model Parameter Dynamics in a Combination Therapy for Bladder Cancer from Sparse Biological Data}

\author{
 Kayode Olumoyin \\
  H. Lee Moffitt Cancer Center and Research Institute\\
  Integrated Mathematical Oncology Department\\
  Tampa, FL \\
  \texttt{kayode.olumoyin@moffitt.org} \\
   \And
 Lamees El Naqa \\
  H. Lee Moffitt Cancer Center and Research Institute\\
  IMO High School Internship Program\\
  Tampa, FL \\
  \texttt{elnaqalamees@gmail.com} \\
  \And
 Katarzyna Rejniak \\
   H. Lee Moffitt Cancer Center and Research Institute\\
  Integrated Mathematical Oncology Department\\
  Tampa, FL \\  
  \texttt{kasia.rejniak@moffitt.org} \\
}

\begin{document}

\maketitle

\begin{abstract}
In a mathematical model of interacting biological organisms, where external interventions may alter behavior over time, traditional models that assume fixed parameters usually do not capture the evolving dynamics. In oncology, this is further exacerbated by the fact that experimental data are often sparse and sometimes are composed of a few time points of tumor volume. In this paper, we propose to learn time-varying interactions between cells, such as those of bladder cancer tumors and immune cells, and their response to a combination of anticancer treatments in a limited data scenario. We employ the physics-informed neural network (PINN) approach to predict possible subpopulation trajectories at time points where no observed data are available. We demonstrate that our approach is consistent with the biological explanation of subpopulation trajectories. Our method provides a framework for learning evolving interactions among biological organisms when external interventions are applied to their environment. 
  
\end{abstract}
%
\section{Introduction}
Bladder cancer is the \(9th\) leading cause of cancer death in men and women globally~\citep{globocan2022}. More than \(50\%\) of patients with intermediate and high-risk localized non-muscle invasive bladder cancer treated with bladder-sparing treatment fail these therapies. In recent years, there is a shift to immunotherapy and combination therapy in how bladder cancer is treated. One of the motivating factor is the success of immunotherapy and combination therapy in the treatment of solid tumors such as unresectable metastatic melanoma and cervical cancer~\citep{Stevanovic2015, pilon2012}. 

Ordinary differential equation (ODE) models can be used to study the interactions between different types of cells in the bladder cancer microenvironment, where external interventions may alter interaction behavior over time. In ~\citep{benaim2024, muller2019}, it is suggested that these external interventions may cause the ODE model parameters to be time-varying. Another motivation for modeling some parameters as time-varying in a biological system is that, in some cases, we lack full mechanistic knowledge of the model. Here, time-varying parameters can act as surrogates for unobserved or unmodeled effects~\citep{glyndavies2025, sawada2022}. Advancements in machine learning and data science have made it possible to learn complex dynamics directly from data. A successful approach in recent years is the physics-informed neural networks (PINNs), introduced in~\citep{raissi2019pinn}, in which the form of the differential equation is assumed to be known and the goal is to learn its parameters from data. Several recent advancements in modeling biological systems from data have been inspired by PINN~\citep{lagergren2020binn, podina2024pinn, kharazmi2021pinn, olumoyin2021pinn}. 

It is commonplace in differential equations model of oncology, that data are sparse. This sparsity occurs because: \((1)\) data are collected at irregular time points; \((2)\) the collected data may contain noise due to measurement errors; and \((3)\) the interacting variables may only be partially observable. In this work, we propose a method that uses the known few discrete data points of the total tumor volume to predict admissible trajectories at unobserved time points for each of the interacting subpopulations. We demonstrate that these learned trajectories at the unobserved time points matches with the biological behavior of the subpopulations. We use spline-based interpolation to extend the observed total tumor volume. We are able to model continuous time-varying parameters from sparse data points.

\section{ODE Model for a Combination Therapy}\label{chemimmu}
We describe an ODE model of a growing tumor exposed to a combination of two anticancer treatments: gemcitabine (GEM) chemotherapy and T cell immunotherapy, based on a study published in \citep{bazargan2023til}. In this experimental study, the \SI{1e5} MB49-OVA bladder cancer cells were instilled orthotopically into the mouse bladder. This was followed on day \(10\) by intravesical injection of \(500 \mu g\) of GEM to locally deplete the tumor microenvironment from the immunosuppressive cells, such as the myeloid-derived suppressive cells (MDSCs).  Next, at day \(14\), the \SI{5e6} OT-1 T cells were administered intravesically into the mouse bladder. The OT-1 T cells are engineered to recognize the MB49-OVA tumor. This immunotherapy treatment is a pre-clinical analogue of the adoptive cell therapy with tumor-infiltrating lymphocytes (ACT-TIL), that is a personalized immunotherapy using patients’ own T cells expanded ex vivo and reinfused to the patient \citep{rosenberg2015til}. In the pre-clinical study, tumor overall volume was monitored using ultrasound imaging and recorded on days \(6, 9, 13, 16, 20\) and \(23\).

We propose the following systems of ODEs to describe the interactions between cancer cells \( (C)\), OT-1 T cells \( (T)\), MDSCs \( (M)\), and GEM \( (G)\). These are presented in Eqs~\eqref{gemot1eqn}, with the summary of the model parameters presented in Table~\eqref{paramtable} in Section~\ref{table_1}. 
\begin{equation}\label{gemot1eqn}
	\begin{split}
		\frac{dC(t)}{dt} &= p_C C(t) - k_{TC} C(t)T(t) - k_{GC} C(t)G(t)\\
		\frac{dT(t)}{dt} &= U_T(t) + n_T T(t) - s_{CT}T(t)C(t) - s_{MT} T(t)M(t) - k_{GT}T(t)G(t)\\
		\frac{dM(t)}{dt} &= r_M C(t) - k_{GM}M(t)G(t) - d_{M} M(t)\\
		\frac{dG(t)}{dt} &= U_G(t)  - d_{G} G(t)
        \end{split}
\end{equation}
Both chemo- and immune-therapy can affect the interactions between cancer cells, OT-1 T cells, and MDSCs. To investigate how the trajectories of these altered dynamics can be inferred from the experimental data, we will focus on the MDSCs suppression of T cells \(s_{MT}\), assuming that this model parameter vary over time due to the external intervention of chemo- and immune-therapy.
\section{Modeling Subpopulations and time-varying Parameters at Unobserved time points using PINN}
 \subsection{Problem Statement}
 Consider a system of ODEs in the following form
 \begin{equation}\label{orig_eqn}
    \frac{du(t)}{dt} = f\!\left(t, u(t); \lambda_1, \lambda_2, \ldots, \lambda_n\right),
\end{equation}
where \(u : [t_0,t_F] \to \mathbb{R}^m\) consists of \(m\) state variables and \(n\) parameters. At most \(p \leq n\) of the parameters are time-varying and denoted by \({\Lambda}(t) = (\lambda_1(t), \lambda_2(t), \ldots, \lambda_p(t))\).  The remaining \((n-p)\) parameters are constant. The function \(f\) is assumed to be nonlinear.
Given experimental data \(\{(t_i, u(t_i))\}_{i=1}^{M^{\text{data}}}\), which are assumed to approximately satisfy Eq.~\eqref{orig_eqn}, we seek to model \(u\) and the time-varying parameters \({\Lambda}\) using neural network surrogates, denoted by \(u_{\text{NN}}\) and \({\Lambda}_{\text{NN}}\), respectively.
 
%
\subsection{PINN Framework and Hyperparameter Tuning}\label{subSEC_mPINN}
 A physics-informed neural network (PINN) was constructed using a feedforward neural network (FNN) as described in Section~\ref{subSEC_FNN}. First, we augment the experimental data points \(\{(t_i, u(t_i))\}_{i=1}^{M^{\text{data}}}\) with an interpolation function \(\hat{u}\) that interpolate \(u(t_i)\) at times \(t_i\), generating \(M^{\text{interp}}\) additional points within the same time domain. 
The PINN takes as input the interpolated time domain. Subpopulation trajectories are approximated through the surrogate \(u_{NN}\), while the interaction dynamics are inferred through \({\Lambda}_{NN}\).  
The \verb+SiLU+ activation function was used in the hidden layers, and the \verb+Softplus+ activation function was applied to the output layers to enforce non-negativity in the solutions. The interpolated experimental data were normalized prior to training and optimization was performed using the \verb+Adam+ optimizer with \(20{,}000\) training epochs to ensure convergence.  
For the other PINN hyperparameters, \(u_{NN}\) was implemented with three hidden layers of 100 neurons each, while \({\Lambda}_{NN}\) was implemented with two hidden layers of 200 neurons each. All implementations were carried out in the \verb+PyTorch+ library.
%
%
\subsection{PINN Training and Loss Function }\label{subSEC_mPINN_2}
The total loss, \(\text{L}_{total}\) (Eq.~\eqref{loss1}), is defined as a weighted sum of four components: the ODE residual loss \(\text{L}_{r}\), the data loss \(\text{L}_{d}\), the initial condition loss \(\text{L}_{IC}\), and the biological constraint loss \(\text{L}_{bc}\). The corresponding weights \(w_{r}, w_{d}, w_{IC}, w_{bc}\) were adaptively adjusted following the multi-task learning of the objective function during training to balance the contributions of the different loss terms~\citep{kendall2018}. The complete formulation of \(\text{L}_{\text{total}}\) is given in Eq.~\eqref{loss2}, where \(\|\cdot\|\) denotes the \(L^2\) norm.  
In Eq.~\eqref{loss2}, the derivative \(\tfrac{d}{dt_j} u_{NN}\) is computed using automatic differentiation as described in~\citep{baydin2018} and evaluated at the time points \(t_j\) where \(j \in \{1,2,\ldots,M^{\text{data}}+M^{\text{interp}}\}\). The data loss \(\text{L}_{d}\) penalizes the difference between \(u_{NN}\) and the experimental data \(u(t_i)\) at times \(t_i\). The residual loss \(\text{L}_{r}\) enforces consistency between the neural network outputs and the governing ODE system, i.e.,
\[
\tfrac{d}{dt_j}u_{NN}(t_j) = f\!\big(t_j, u_{NN}(t_j); {\Lambda}_{NN}(t_j), {\lambda_{NN}}_{p+1}, \ldots, {\lambda_{NN}}_{n} \big).
\]
The initial condition loss \(\text{L}_{IC}\) ensures that the learned trajectories satisfy prescribed initial conditions, thereby improving stability during training. Finally, the biological constraint loss \(\text{L}_{bc}\) incorporates domain-specific constraints on the subpopulations at \(M^{\text{bc}}\) specified time points.
%
%
{\small
\begin{equation}\label{loss1}
	\begin{split}
		\text{L}_{total} = w_{r} \text{L}_{r} + w_{d} \text{L}_{d} + w_{IC} \text{L}_{IC} +  w_{bc} \text{L}_{bc}
        \end{split}
\end{equation}
}
{\small
\begin{equation}\label{loss2}
\begin{aligned}
\text{L}_{r} &= \tfrac{1}{M^{\text{data}}+M^{\text{interp}}}\!\sum_{j=1}^{M^{\text{data}}+M^{\text{interp}}}\!
  \Big\|\tfrac{d}{dt_j}u_{NN}(t_j)-f\!\big(t_j,u_{NN}(t_j);{\Lambda}_{NN}(t_j),{\lambda_{NN}}_{p+1},\ldots,{\lambda_{NN}}_{n}\big)\Big\|, \\
\text{L}_{d} &= \tfrac{1}{M^{\text{data}}}\!\sum_{i=1}^{M^{\text{data}}}\!\|u_{NN}(t_i)-\hat{u}(t_i)\|, \quad
\text{L}_{IC} = \|u_{NN}(t_0)-\hat{u}(t_0)\|, \quad
\text{L}_{bc} = \tfrac{1}{M^{\text{bc}}}\!\sum_{k=1}^{M^{\text{bc}}}\!\|u_{NN}(t_k)-\hat{u}(t_k)\|.
\end{aligned}
\end{equation}
}
\section{Result}
In our PINN implementation, we assumed the total tumor volume to be the sum of the subpopulations \( C\), \( T\), and \( M\), which are unknown at the sparse experimental time points. We increased the number of data points within the points where the experimental data points were collected using spline-based interpolation. 

\begin{figure}[h!]
    \centering
    \begin{subfigure}[b]{0.9990\textwidth}
        \centering
        \includegraphics[width=\textwidth]{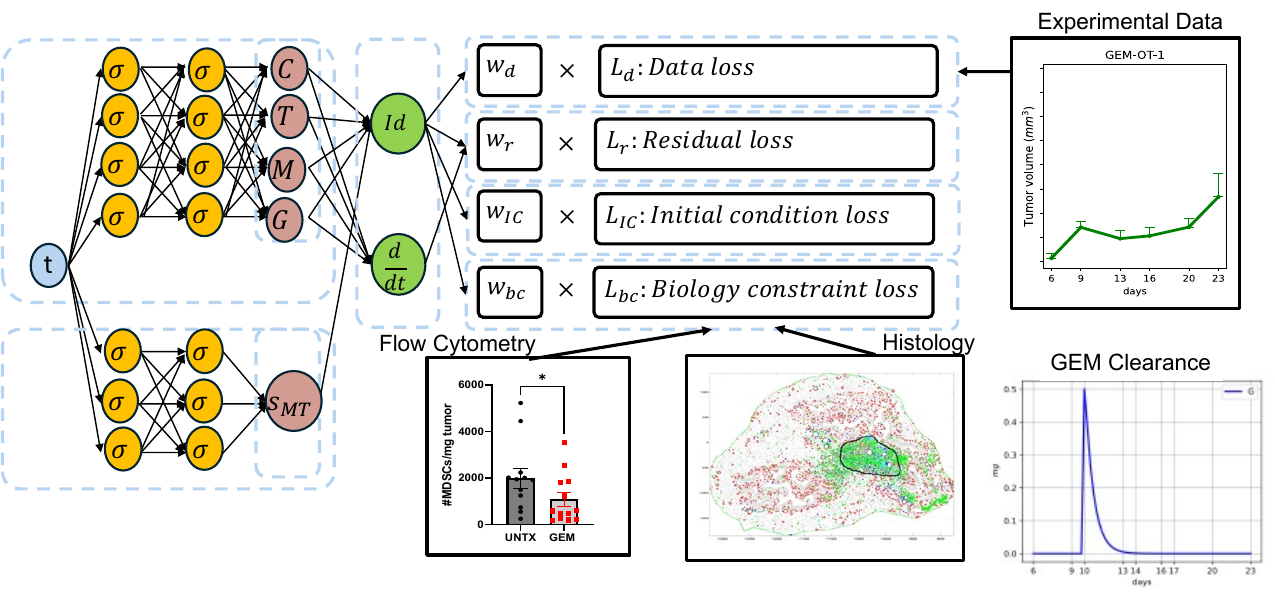}
        \caption{PINN framework to learn subpopulations and a time-varying parameter.}
        \label{fig:wide1}
    \end{subfigure}
    \vspace{0.3cm}
    \begin{subfigure}[b]{0.990\textwidth}
        \centering
        \includegraphics[width=\textwidth]{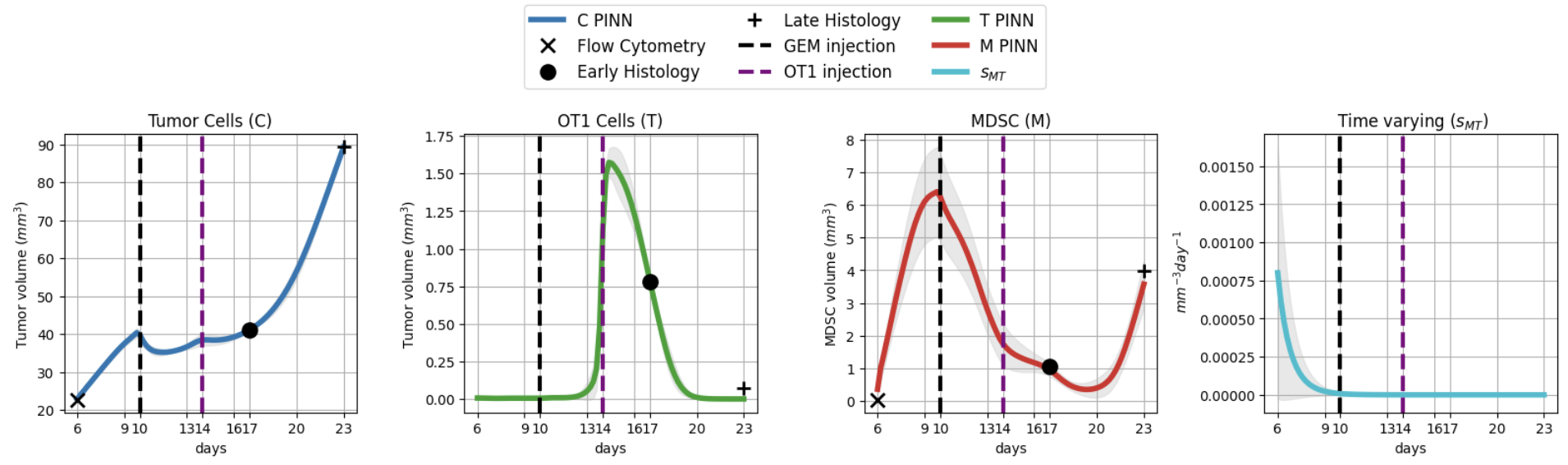}
        \caption{Subpopulations \(C, T, M \) and time-varying \(s_{MT}\) learned by the PINN.}
        \label{fig:wide2}
    \end{subfigure}
    \caption{PINN implementation to the combination therapy in Eq.~\eqref{gemot1eqn}.}
    \label{expdata}
\end{figure}

%
We denote the experimental data as \(u_{\text{GEM-OT1}}\), which are tumor volume measurements from ultrasound images recorded on days \(6, 9, 13, 16, 20\), and \(23\), see Figure~\eqref{expdata}({\bf a}).  We obtained the proportions of \(C\) and \(M\) at \(t_0 = 6\) from flow cytometry data (Figure~\eqref{expdata}({\bf a})). Using two histology time points data collected on days \(17\) and \(23\), we obtained digital histology of bladder tissue from a mouse treated with GEM and OT-1 T cells. In the digital histology presented in Figure~\eqref{expdata}({\bf a}), tumors \((C)\) are outlined in black, MDSCs \((M)\) are represented as green dots, and T cells \((T)\) are blue dots. See Section~\ref{chemimmu2} for how we set the initial conditions and the biology constraints in the PINN implementation of the combination therapy in Eq.~\eqref{gemot1eqn}.

We demonstrate our approach by applying the PINN framework and implementation (sections~\eqref{subSEC_mPINN} and \eqref{subSEC_mPINN_2}) to an ODEs model of a growing tumor exposed to a combination of two anticancer treatments: gemcitabine (GEM) chemotherapy and T cell immunotherapy described in section~\eqref{chemimmu}. In Figure~\eqref{expdata}({\bf a}), we present the schematic of the implementation of the PINN algorithm to the system of ODEs given in Eq.~\eqref{gemot1eqn}. First, PINN learns the dynamics of the subpopulations \( C\), \( T\), \( M\), and \( G\), which are potential solutions to Eq.~\eqref{gemot1eqn}. Next, in an unsupervised way, PINN learns the time-varying parameter \(s_{MT}\) to infer interaction dynamics resulting from the external interventions of GEM (chemotherapy) and OT-1 T cells (immunotherapy). 

%
As shown in Figures~\eqref{expdata}({\bf b}), PINN learns a possible dynamics for the subpopulations \( C\), \( T\), and \( M\), aligning with biological constraints at the flow cytometry point (represented as black cross) and histology time points (represented as black circle dot at the early histology time point and black plus at the late histology time point). At the GEM injection (black vertical dash line) we see the effect of local depletion of the MDSCs and some killing of the tumor cells. The OT-1 injection (purple vertical dash line) indicate some killing of the tumor cells. The learned clearance of the drug GEM \((G)\) is shown in Figure~\eqref{expdata}({\bf a}). In Figures~\eqref{expdata}({\bf b}), PINN learns possible interaction dynamics for the model parameter \(s_{MT}\) in an unsupervised fashion. We ran the PINN algorithm \(10\) times to generate uncertainty bound for each of the PINN solutions. The sum of the Subpopulations \(C, T, M \) matches the experimental tumor volume data at the time points where measurements were collected (Section~\ref{ctm_G}).
\section{Discussion}
The results we present in this work captures the changing dynamics in a limited-data scenario involving three different types of interacting cells and drug Gemcitabine. Additionally, we demonstrate how to use the PINN framework to incorporate biological knowledge (histology and flow cytometry data) as a form of regularization. This enables us to reduce the space of admissible solutions. The approach is applicable to many other differential equation models in biology, particularly in systems where interacting organisms alter their behavior in response to external interventions.
\begin{ack}
This work was supported by the Department of Defense grant W81XWH-22-1-0340 and the US National Institutes of Health, National Cancer Institute grant R01-CA259387. This work was supported in part by the Shared Resources at the H. Lee Moffitt Cancer Center \& Research Institute an NCI designated Comprehensive Cancer Center under the grant P30-CA076292 from the National Institutes of Health. LEN was supported by the NIH grant U54 CA274507-02A1 through the Moffitt High-School Internship Program in Integrated Mathematical Oncology (HIP-IMO).
%
%
%
%
\end{ack}

\bibliographystyle{abbrv}
\bibliography{references}

\newpage

\appendix
\section{Technical Appendices and Supplementary Material}
%
\subsection{Table of parameters for the ODE in Eq.~\eqref{gemot1eqn}}\label{table_1}
\begin{table}[h]
  \caption{Model parameters in Eq.~\eqref{gemot1eqn}}
  \label{paramtable}
  \centering
  \resizebox{0.99\textwidth}{!}{%
  \begin{tabular}{ll|ll}
    \toprule
    Parameter & Description [units] & Parameter & Description [units] \\
    \midrule
    $k_{TC}$ & T cell killing of cancer [$mm^{-3} \, day^{-1}$] & $n_{T}$  & T cell net growth [$day^{-1}$] \\
    $k_{GC}$ & GEM killing of cancer [$mg^{-3} \, day^{-1}$]    & $U_{G}$  & GEM injection [$mg \, day^{-1}$] \\
    $k_{GM}$ & GEM killing of MDSCs [$mg^{-3} \, day^{-1}$]     & $U_{T}$  & T cell injection [$mm^{3} \, day^{-1}$] \\
    $k_{GT}$ & GEM killing of T cells [$mg^{-3} \, day^{-1}$]   & $r_{M}$  & MDSCs recruitment [$day^{-1}$] \\
    $d_{G}$  & GEM clearance [$day^{-1}$]                   & $s_{MT}$ & MDSCs suppression of T cells [$mm^{-3} \, day^{-1}$] \\
    $d_{M}$  & MDSCs death [$day^{-1}$]                      & $s_{CT}$ & Cancer suppression of T cells [$mm^{-3} \, day^{-1}$] \\
    $p_{C}$  & Cancer net growth [$day^{-1}$]              & \\
    \bottomrule
  \end{tabular}}
\end{table}
\subsection{Feedforward Neural Network (FNN)} \label{subSEC_FNN}
An FNN is an artificial neural network where the forward pass is in one direction, and it is usually fully connected. It can be represented as a composition of functions with \(L\) layers, where the input layer is an affine transformation of a vector \(t\) followed by an activation function. The output of the input layer becomes input to the next layer, continuing until the \(L\)th layer as shown in Eq.~\eqref{fnneq}.  
{\small
\begin{equation}\label{fnneq}
 NN(t;\theta) = \sigma_{L}(W_{L}\sigma_{L-1}(W_{L-1}\ldots \sigma_{2}(W_{2} \sigma_{1}(W_{1}t + b_1)+b_2)\ldots+b_{L-1})+b_L),
 \end{equation}
 }
 \noindent where \(\theta : = (W_1, \ldots, W_L,b_1, \ldots, b_L)\) denotes the set of neural network weight matrices \(W_k\) and bias vectors \(b_k\) for \((k = 1, \ldots, L)\), and \(\sigma_k (\cdot)\) denotes the activation function. It is widely known that a feedforward neural network with at least one hidden layer containing a finite number of neurons can approximate any continuous function on compact subsets of \({\rm I\!R}^m\)~\citep{hornik1991}. 
\subsection{Setting the Initial Conditions and Biology Constraints for PINN}\label{chemimmu2}
 We assume that \(u_{\text{GEM-OT1}}\) is a sum of the subpopulations \( C\), \( T\), and \( M\), so we implement the data loss \(\text{L}_{d}\) in Eq.~\eqref{loss2} as \( \frac{1}{M^{\text{data}}}\sum_{i = 1} ^ {M^{\text{data}}} ((C(t_i) + T(t_i) + M(t_i)) - u_{\text{GEM-OT1}}(t_i)  )^{2}  \). In order to initialize \( C\), \( T\), \( M\), at \(t_0 = 6\), we define \(\text{L}_{IC}\) in Eq.~\eqref{loss2} as follows:  
{\small
\[
\text{L}_{IC} = \left( C(t_0) - q_0 \times u_{\text{GEM-OT1}}(t_0)  \right)^{2}  +  \left( T(t_0) - q_1 \times u_{\text{GEM-OT1}}(t_0)  \right)^{2} +  \left( M(t_0) - q_2 \times u_{\text{GEM-OT1}}(t_0)  \right)^{2},
\]
}
where \(q_0 = 0.99887\), \(q_1 = 0\), and \(q_2 = 1 - 0.99887\).

Similarly, we obtain the biology constraints loss \(\text{L}_{bc}\) in Eq.~\eqref{loss2} as follows:
{\small
\[
\text{L}_{bc} = \sum_{k \in \{ 1,2\}}\left(  C(t_{k}) - q_{0_{k}} \times u_{\text{GEM-OT1}}(t_{k}) \right) ^{2}  + \left(  T(t_{k}) - q_{1_{k}} \times u_{\text{GEM-OT1}}(t_{k}) \right) ^{2}  + \left(  M(t_{k}) - q_{2_{k}} \times u_{\text{GEM-OT1}}(t_{k}) \right) ^{2} 
\]
}
Where \(q_{0_{1}} = 0.95755\), \(q_{1_{1}} = 0.01818\), \(q_{2_{1}} = 0.02427\) are the proportions of \( C\), \( T\), and \( M\) respectively at day \(17\) (histology time point).  Similarly for \( C\), \( T\), and \( M\) at day \(23\) (histology time point), we have that \(q_{0_{2}} = 0.95665\), \(q_{1_{2}} = 0.00078\), and \(q_{2_{2}} = 0.04256\) respectively.
\subsection{Loss Convergence, Comparison of learned Subpopulations with Experimental data}\label{ctm_G}
The loss convergence and the comparison of the sum of the subpopulations \( C+T+M\) vs the experimental tumor volume data at the time points where measurements were collected.
\begin{figure}[htbp]
    \centering
    \begin{subfigure}[b]{0.450\textwidth}
        \centering
        \includegraphics[width=\textwidth]{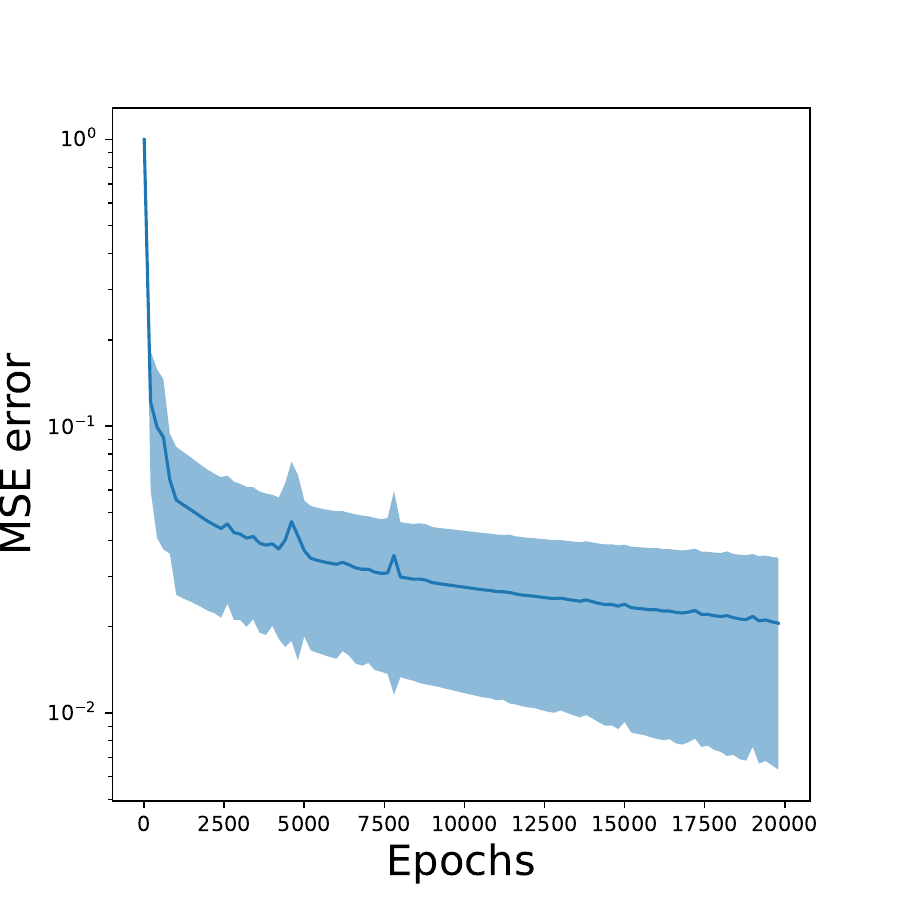}
        \caption{Convergence plot for the PINN implementation.}
        \label{fig:wide3}
    \end{subfigure}
    \hspace{0.15cm}
    \begin{subfigure}[b]{0.525\textwidth}
        \centering
        \includegraphics[width=\textwidth]{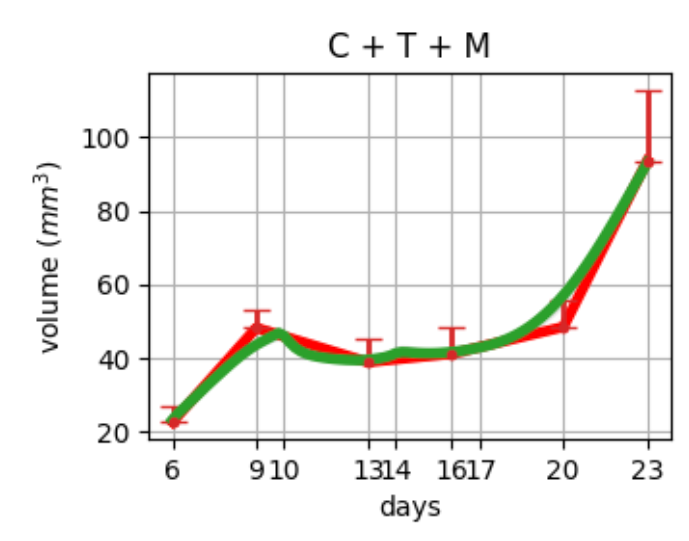}
        \caption{Total tumor volume (red) vs learned \(C+T+M \) (green).}
        \label{fig:wide4}
    \end{subfigure}
    \label{expdata2}
\end{figure}
%

%
%
 

\end{document}